\documentclass[letterpaper, 10 pt, conference]{support/ieeeconf}
\usepackage{mathtools}
\usepackage{times}
\usepackage[pdftex]{graphicx}
\usepackage{subfigure}
\usepackage{amsmath,amssymb,amsopn,amstext,amsfonts}
\usepackage{cancel}
\usepackage[space]{cite}
\usepackage{pdfsync}
\usepackage{balance}
\usepackage{color}
\usepackage[ruled,vlined]{algorithm2e}
\usepackage{bm}

\usepackage{diagbox}
\usepackage{float}
\usepackage{epstopdf}
\usepackage{pifont}
\usepackage{fixltx2e}
\usepackage{amsmath}
\usepackage{multirow}
\usepackage{booktabs}
\usepackage{url}
\usepackage{svg}
\usepackage{bm}
\usepackage{threeparttable}
\usepackage[linkcolor=black,citecolor=black,urlcolor=black,colorlinks=true]{hyperref}
\usepackage{siunitx,array}
\usepackage{lipsum}
\usepackage{makecell}
\usepackage{caption}
\usepackage{mathrsfs}

\newcommand{\tp}{^{\mathrm{T}}}

\makeatletter
\newcommand{\setword}[2]{%
  \phantomsection
  #1\def\@currentlabel{\unexpanded{#1}}\label{#2}%
}
\makeatother

\graphicspath{{./figures/}}
\DeclareGraphicsExtensions{.png,.jpg,.eps,.pdf}
\IEEEoverridecommandlockouts
\makeatletter
\def\endthebibliography{%
	\def\@noitemerr{\@latex@warning{Empty `thebibliography' environment}}%
	\endlist
}

\title{\textbf{Polynomial-based Online Planning for\\
		Autonomous Drone Racing in Dynamic Environments}}
\author{
        Qianhao Wang $^{\dag}$, 
        Dong Wang $^{\dag}$, 
        Chao Xu,
        Alan Gao,
        and Fei Gao
	    \thanks{\textbf{${\dag}$ Equal contribution.}}        
        \thanks{All authors are with the College of Control Science and Engineering, Zhejiang University, Hangzhou, 310027, China, and also with the Huzhou Institute of Zhejiang University, Huzhou, 313000, China. }
	    \thanks{Email:{\tt\small \{qhwangaa, fgaoaa\}@zju.edu.cn}} 
	    \thanks{Corresponding Author: Fei Gao.}}  

\begin{document}

\maketitle
\thispagestyle{empty}
\pagestyle{empty}

\begin{abstract}
    In recent years, there is a noteworthy advancement in autonomous drone racing. However, the primary focus is on attaining execution times, while scant attention is given to the challenges of dynamic environments. The high-speed nature of racing scenarios, coupled with the potential for unforeseeable environmental alterations, present stringent requirements for online replanning and its timeliness.
    For racing in dynamic environments, we propose an online replanning framework with an efficient polynomial trajectory representation.
    We trade off between aggressive speed and flexible obstacle avoidance based on an optimization approach.
    Additionally, to ensure safety and precision when crossing intermediate racing waypoints, we formulate the demand as hard constraints during planning.
    For dynamic obstacles, parallel multi-topology trajectory planning is designed based on engineering considerations to prevent racing time loss due to local optimums.
    The framework is integrated into a quadrotor system and successfully demonstrated at the DJI Robomaster Intelligent UAV Championship, where it successfully complete the racing track and placed first,  finishing in less than half the time of the second-place\footnote{\label{foot_rankings}https://pro-robomasters-hz-n5i3.oss-cn-hangzhou.aliyuncs.com/sass/event-list.html}.
\end{abstract}

\section{Introduction}
\label{sec:Introduction}
Quadrotors are gaining popularity in various industrial and commercial scenarios due to their versatility and exceptional performance.
In recent years, autonomous drone racing, a research field focusing on planning trajectories for quadrotors to follow an aggressive reference routine while precisely crossing some intermediate landmarks, receives considerable attentions~\cite{romero2022time,han2021fast,foehn2021time,hanover2023autonomous} and sparkes an international competition craze, such as the AlphaPilot Challenge \cite{foehn2022alphapilot,guerra2019flightgoggles} and the Autonomous Drone Race~\cite{moon2019challenges,cocoma2019towards} in IEEE IROS.
The ultimate pursuit of minimizing the execution time in drone racing increasingly ignites the fire for quadrotors to be applied in several emergencies, such as post-disaster communications and urgent transportation of essential supplies.

In such scenarios, dynamic obstacles or moving landmarks that require investigation and traversal will inevitably arise due to environmental changes.
How to plan a minimum-time trajectory through a series of waypoints in dynamic environments remains a challenging problem that can not be completely solved by previous works.
In detail, this problem requires that the planning method satisfies the conditions simultaneously: 
\setword{\textbf{(1)}}{requirement1}~highest possible speed of completing the track;
\setword{\textbf{(2)}}{requirement2}~agile avoidance of the dynamic obstacles;
\setword{\textbf{(3)}}{requirement3}~shortest possible computational time of trajectory replan;
\setword{\textbf{(4)}}{requirement4}~precise traversal of the waypoints.

\begin{figure}[!t]
	\centering
    \vspace{0.2cm}
	\includegraphics[width=1\linewidth]{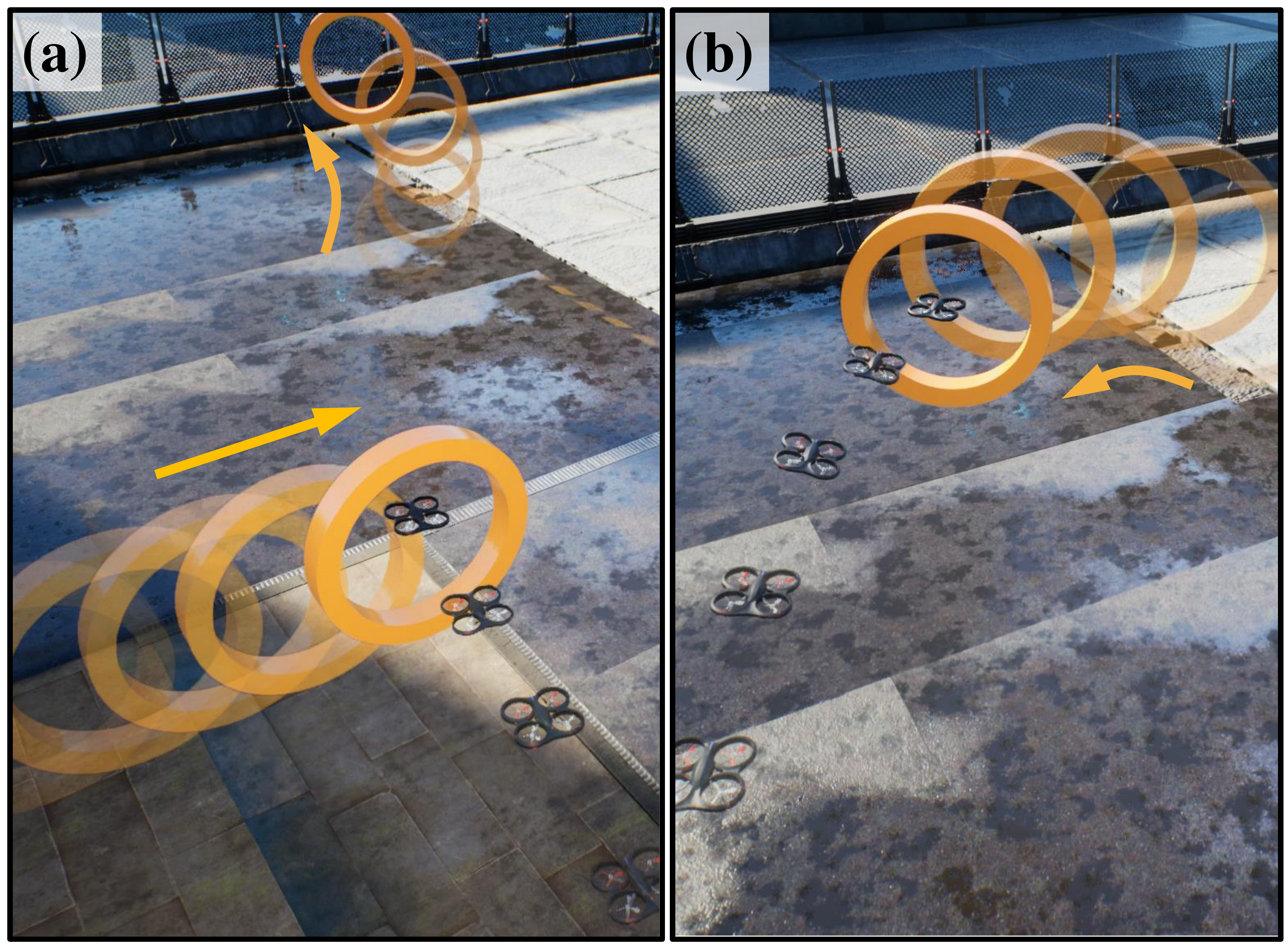}
    \captionsetup{font={footnotesize}}
	\caption{
        The snapshot of our online planning method applied in a challenging dynamic environment, where the quadrotor is required to precisely
        pass through two dynamic gates in order.
        In \textbf{(a)}, the quadrotor passes through the first gate and flies towards the second one.
        \textbf{(b)} depicts a close-up of crossing the second gate.
	}
	\label{fig:toutu}
    \vspace{-1.0cm}
\end{figure}

The first two points \ref{requirement1} and \ref{requirement2} are intuitive for safe racing in dynamic environments.
As for \ref{requirement3}, since not all variations in the dynamic surroundings can be preemptively observed and accurately predicted, these unforeseeable circumstances require online replanning.
Particularly during high-speed drone racing, the efficiency of replanning is crucial for ensuring the timeliness of trajectories. 
Although existing works~\cite{foehn2021time,romero2022time} demonstrate the ability to compete with skilled human pilots in racing, minute-level computational demands make them prohibitive to respond to unforeseeable changes.
For \ref{requirement4}, as shown in Fig.~\ref{fig:toutu}, some waypoints may have low spatial tolerance even in motion.
Inaccurate traversal can cause collision or mission failure.
However, most racing works~\cite{han2021fast,romero2022model} do not guarantee stable and precise traversal in planning, because of the use of soft constraint approach that relies on parameters.


In order to address the aforementioned requirements, we propose a strong polynomial-based online planning framework for racing in dynamic environments by incorporating careful engineering considerations with our previous works.
We obtain the trajectory by an optimization method.
For online replanning of \ref{requirement3}, we adopt MINCO~\cite{wang2022geometrically} as the trajectory representation and improve it to a time-uniform version.
This implementation is more lightweight yet still maintains the capacity to allow for spatial-temporal deformations, improving computational efficiency. 
For rapid speed of \ref{requirement1}, we boost the aggressiveness of quadrotors by minimizing execution time in the optimization based on the temporal freedom of trajectory.
For precise waypoint traversal of \ref{requirement4}, we formulate this waypoint-through requirement into a hard constraint to ensure safety and stability, regardless of whether the waypoint is static or in motion.
For obstacle avoidance of \ref{requirement2}, we build upon our previous work~\cite{wang2021autonomous} by calculating multiple trajectories in parallel under different topologies segmented by dynamic obstacles and then selecting the optimal one.


Finally, we integrate the proposed planning framework into a customized quadrotor system, combining state estimation, control, and real-time vision-based detection modules. 
This system was deployed at the 2022 \textbf{DJI Robomaster Intelligent UAV Championship}\footnote{\label{dji_bisai_web}https://www.robomaster.com/zh-CN/robo/drone}, where quadrotors are tasked with navigating a track with dynamic obstacles, narrow gaps that require SE(3) planning, and gates.
The gates must be traversed in a specific order as fast as possible, even though they are either in motion or have a random location within a range.
In this competition, our system succeeded in completing the racing track and placed first, which proves our method's comprehensive capabilities of performing high-speed flight in challenging dynamic environment.

To summarize, the contributions of this paper are:
\begin{itemize}
    \item We implement time-uniform MINCO by improving our previous work, which boosts computational efficiency to enhance the timeliness of replan for high-speed flight. 
    \item We achieve a hard constraint on the position of drone during crossing both static and moving waypoints, to ensure precision of crossing, by modeling the position of waypoints as trajectories about time and formulating them as boundary conditions for joint optimization.
    \item We propose a replanning framework, combined with evaluating different topologies segmented by dynamic obstacles and large attitude flight, to deal with challenging environments.
    Ablation experiments and competition prove our method's effectiveness for racing in dynamic environments.
\end{itemize}

\section{Related Work}
\label{sec:relatedworks}
We divide racing works online and offline depending on whether the method can perform in real-time.

In general, offline methods are more comprehensive in problem construction and can serve as the baseline for online methods.
Recently, there is a remarkable work\cite{foehn2021time} of Foehn et al. to generate time-optimal trajectories for drones to outperform professional pilots in racing.
They use discrete state points to represent the trajectory and solve it using an optimization method while formulating the gate-through requirement as a complementary constraint.
Additionally, they adopt a full-state quadrotor model with the thrust of single rotor as control input and impose constraint on each rotor, which saturates the actuator to achieve optimal time.
Different from traditional racing, Han et al.\cite{han2021fast} focus on planning SE(3) trajectories to cross narrow gaps.
They opt for polynomial trajectory and use the differential flatness of the quadrotor~\cite{mellinger2011minimum} to deliver spatial constraints to SE(3) state.
Compared to solving a large-scale problem in Wang's work\cite{wang2022geometrically}, this method develops parallel computing for this planning problem, significantly enhancing efficiency.
As for cluttered environments, Penicka et al.\cite{penicka2022minimum} extend on Li's work\cite{li2016asymptotically} using a hierarchical sampling-based framework guided with an incrementally more complex quadrotor model.
However, above methods do not have a real-time performance, making them inadequate to adapt to unpredictable changes, such as disturbances in the position of gate.

To enable the drone to handle environmental changes, online replanning is essential.
Some search-based methods\cite{liu2017search,liu2018search} add the need to bring time to a minimum in the cost function to achieve real-time replan for rapid flight.
But they can merely limit the acceleration of every axis while failing to explore the boundaries of the actuator, leading to conservative practices.
For the 2019 AlphaPliot Challenge, Foehn\cite{foehn2022alphapilot} perform an onboard detection of the gates as a reference for replanning.
They generate motion primitives based on maximum acceleration and obtain the time-optimal trajectory by sampling velocity states.
Finally, they use a polynomial to fit the trajectory to track.
Afterward, Romero~\cite{romero2022time} replace the polynomial parameterization of the above work\cite{foehn2022alphapilot} and use the Model Predictive Contouring Control (MPCC)\cite{romero2022model} considering an accurate full-state model which is extended to include a linear drag model\cite{faessler2017differential} to execute the trajectory obtained from sampling.
Due to the penalty on progress item and constraining the dynamics of each individual rotor in MPCC, this approach can better exploit the drone's performance limits.
Nevertheless, both methods of leaving it to the controller to track the trajectory that ignores the dynamic feasibility constraints can only ensure the safety of the gate-through by adding a certain cost weight at the controller side.
This is rarely applicable to dynamic environments.

\begin{table*}[!t]
    \centering
    \caption{Comparison of Different Trajectory Representations}
    \label{tab:parameter}
    \small
    \begin{tabular*}{\textwidth}{
      @{\extracolsep{\fill}}
      m{1.4cm}
      >{$\displaystyle}l<{\vphantom{\sum_{1}{N}}$}
      >{\refstepcounter{equation}(\theequation)}r
      >{$\displaystyle}l<{\vphantom{\sum_{1}{N}}$}
      >{\refstepcounter{equation}(\theequation)}r
      >{$\displaystyle}l<{\vphantom{\sum_{1}{N}}$}
      >{\refstepcounter{equation}(\theequation)}r
      @{}
    }
    \toprule
      & \multicolumn{2}{l}{MINCO} & \multicolumn{2}{l}{Time-uniform MINCO}  & \multicolumn{2}{l}{Normalized Time-uniform MINCO} \\
    \midrule
        \makecell[l]{time \\allocation} &
        \mathbf{T} = (T_1,\dots,T_{M})\tp, &
        \label{eq:minco_T} & 
        \hat{\mathbf{T}} = (T/M)\mathbf1, &
        \label{eq:u_minco_T} & 
        \bar{\mathbf{T}} = \mathbf1, &
        \label{eq:n_u_minco_T}
        \\

        boundary conditions &
        \mathbf{z}^o, \mathbf{z}^f,&
        \label{eq:minco_b} & 
        \mathbf{z}^o, \mathbf{z}^f,&
        \label{eq:u_minco_b} & 
        \makecell[l]{\bar{\mathbf{z}}^o = \mathbf{S}_{s}(T/M){\mathbf{z}}^o,\\\bar{\mathbf{z}}^f = \mathbf{S}_{s}(T/M){\mathbf{z}}^f,} &
        \label{eq:n_u_minco_b}
        \\

        mapping equation &
        \mathbf{M}({\mathbf{T}}){\mathbf{C}} = \mathbf{b}(\mathbf{Q},~\mathbf{z}^o,\mathbf{z}^f), & 
        \label{eq:minco_mapping} & 
        \mathbf{M}(\hat{\mathbf{T}})\hat{\mathbf{C}} = \mathbf{b}(\mathbf{Q},~\mathbf{z}^o,\mathbf{z}^f), &
        \label{eq:u_minco_mapping} & 
        \mathbf{M}(\mathbf{1})\bar{\mathbf{C}} = \mathbf{b}(\mathbf{Q},~\bar{\mathbf{z}}^o, \bar{\mathbf{z}}^f), &
        \label{eq:n_u_minco_mapping}
        \\
        
        coefficients solving &
        \multicolumn{2}{l}{
            \makecell[l]{by online PLU factorization and \\
            solving linear systems of equations}    
        } &
        \hat{\mathbf{c}}_i = \mathbf{S}_{2s}(M/T)\bar{\mathbf{c}}_i,&
        \label{eq:u_minco_c} & 
        \bar{\mathbf{C}} = \mathbf{M}^{-1}(\mathbf{1})\mathbf{b}(\mathbf{Q},~\bar{\mathbf{z}}^o, \bar{\mathbf{z}}^f).&
        \label{eq:n_u_minco_c}
        \\

    \bottomrule
    \end{tabular*}
    \vspace*{-0.4cm}
\end{table*}

As for the navigation works mentioned for collision-free flight, they concentrate on handling the information about the static or dynamic environment\cite{wang2021autonomous,chen2022rast} obtained from perception to serve subsequent trajectory generation and on constructing constraints to avoid obstacles\cite{zhou2020ego,zhou2021raptor}.
However, they have a shortage of adaptability for drone racing tasks, which require the quadrotor to traverse waypoints in order with a focus on execution time.


\section{Trajectory Representation}
\label{sec:traj}



In this section, for enhancing the computational efficiency for online replanning, we implement time-uniform MINCO based on our previous work~\cite{wang2022geometrically} to conduct spatial-temporal deformation of the flat-output trajectory.
We present the definitions and comparison in Tab.~\ref{tab:parameter}.
Essentially time-uniform MINCO and the normalized one are just two special kinds of MINCO.
Therefore, to distinguish the special parts of the different trajectory representations, we use the symbols with hat $\hat{\cdot}$ and the symbols with horizontal lines $\bar{\cdot}$, such as Eq.(\ref{eq:u_minco_T}) and Eq.(\ref{eq:n_u_minco_T}), to denote the special parts in the time-uniform MINCO and normalized one, respectively.
In the following, we go over the meaning of the variables that appear in Tab.~\ref{tab:parameter}.

An $s$-order MINCO is defined as an $m$-dimensional $M$-piece polynomial trajectory. The $i$-th piece trajectory is defined by an $\mathcal{D}=2s-1$ degree polynomial as
\begin{equation}
    \label{eq:minco_def}
        p_i(t) = \mathbf{c}_i\tp\beta(t),~~\forall t \in [0, T_i],
\end{equation}
where $\beta(x):=(1,x,\dots,x^{\mathcal{D}})\tp$ is the natural basis.
$\mathbf{c}_i$  and $T_i$ means the coefficients and duration of the $i$-th piece respectively.
For the $M$-piece trajectory, the total duration is $T=\sum_{i=1}^{M}T_i$, and its coefficients $\mathbf{C}$, time allocation $\mathbf{T}$ and intermediate points $\mathbf{Q}$ can be written as
\begin{equation}
\label{eq:seg_parameter}
\begin{aligned}
    \mathbf{C} &= (\mathbf{c}_1\tp,\dots,\mathbf{c}_i\tp,\dots,\mathbf{c}_{M}\tp)\tp &&\in \mathbb{R}^{2Ms \times m},\\
    \mathbf{T} &= (T_1,\dots,{T}_i,\dots,T_{M})\tp &&\in \mathbb{R}_{>0}^{M},\\
    \mathbf{Q} &= (\mathbf{q}_1,\dots,\mathbf{q}_i,\dots,\mathbf{q}_{M-1}) &&\in \mathbb{R}^{m\times(M-1)},
\end{aligned}
\end{equation}
where $\mathbf{q}_i$ means the $0$-order derivative of $p_i(t)$ at $T_i$.
And in Tab.~\ref{tab:parameter}, $\mathbf{z}^o, \mathbf{z}^f \in \mathbb{R}^{m\times s}$ is the boundary conditions containing high order derivative at $p_1(0)$ and $p_M(T_M)$.
$ \mathbf{M} \in \mathbb{R}^{2Ms\times 2Ms}$ and $ \mathbf{b} \in \mathbb{R}^{2Ms\times m}$ are matrixes with $\mathbf{T}$ and $\mathbf{Q},~\mathbf{z}^o,\mathbf{z}^f$ as variables, respectively, which refer to the Eq.(54,~55) in \cite{wang2022geometrically} for the detailed definition. 
In Eq.(\ref{eq:u_minco_T},~\ref{eq:n_u_minco_T}), $\mathbf{1}=\left(1,1,...,1\right)\tp \in \mathbb{R}^{M}$.
In Eq.(\ref{eq:n_u_minco_b},~\ref{eq:u_minco_c}), $\mathbf{S}_{y}(x):=$ Diag$(1,x,\dots,x^{y-1})$.



\subsection{MINCO Trajectory}
\label{subsec:minco}

For an $s$-order MINCO, given boundary conditions Eq.(\ref{eq:minco_b}), intermediate points $\mathbf{Q}$ and time allocation Eq.(\ref{eq:minco_T}), the coefficients ${\mathbf{C}}$ can be uniquely determined by $\mathbf{z}^o,\mathbf{z}^f,\mathbf{Q}$ and ${\mathbf{T}}$ based on the linear mapping equation Eq.(\ref{eq:minco_mapping}), which is defined as Theorem 2 in \cite{wang2022geometrically}.
Because $\mathbf{M}$ is a nonsingular banded matrix, based on its banded PLU factorization, the coefficients can be computed by solving two linear systems of equations with linear time and space complexity~\cite{golub2013matrix}, which prevents the need to explicitly calculate $\mathbf{M}^{-1}$.

For user-defined penalty function $F({\mathbf{C}},{\mathbf{T}})$ with available gradients, MINCO serves as a linear-complexity differentiable layer $H(\mathbf{z}^o,\mathbf{z}^f,\mathbf{Q},{\mathbf{T}})=F({\mathbf{C}},{\mathbf{T}})$.
To accomplish the deformation of MINCO, we need to obtain the gradients of $H$ w.r.t. the trajectory's variable $\mathbf{z}^o,\mathbf{z}^f,\mathbf{Q}$ and ${\mathbf{T}}$ from the given gradients $\partial{F}/\partial{{\mathbf{C}}}$ and $\partial{F}/\partial{{\mathbf{T}}}$ as  Eq.(60,~68) in \cite{wang2022geometrically}.
During the process, results of right multiplying $\mathbf{M}^{-1}$ by a vector are needed several times. 
Similar to coefficients solving, these results can be get by using the PLU factorization of $\mathbf{M}$.

\subsection{Time-uniform MINCO Trajectory}
It should be emphasized that, in Tab.~\ref{tab:parameter} and this section,  the role of normalized time-uniform MINCO is to serve as an intermediate state of the time-uniform MINCO to aid in the calculation of coefficients and gradients.

  
We refer to a special MINCO trajectory with uniform time allocation Eq.(\ref{eq:u_minco_T}) as time-uniform MINCO, since each piece has the same duration.
The time-uniform MINCO still allows for spatial-temporal deformation, which is accomplished through the freedom of the total time $T$.

As Eq.(\ref{eq:u_minco_mapping}) states, even with a uniform time allocation, we still need to online deal with $\mathbf{M}^{-1}$ whenever $T$ changes.
To avoid PLU factorization of $\mathbf{M}$ every time, using temporal scaling, we define normalized time-uniform MINCO $\bar{p}(t)$ of a time-uniform MINCO $\hat{p}(t)$, and the $i$-th piece is
\begin{equation}
    \label{eq:time_uniform_minco}
    \bar{p}_i(t) = \hat{p}_i({T}/{M} \cdot t),~~\forall t \in [0, 1],
\end{equation}
where $\bar{p}_i(t)$ takes $1$ as piece duration.
Since the original trajectory is time-uniform,
As defined in Eq.(\ref{eq:time_uniform_minco}), the intermediate waypoints $\mathbf{Q}$ do not change because the spatial shape of the trajectory is kept constant.
The high order derivatives of $\bar{p}_i(t)$ can be written as
\begin{equation}
    \bar{p}_i^{(s)}(t) = (T/M)^s \hat{p}_i^{(s)}({T}/{M} \cdot t),~~\forall t \in [0, 1].
\end{equation}
Then the boundary condition is deflated by a factor of $(T/M)^s$ in the $s$-th order derivative due to the temporal scaling, as written in Eq.(\ref{eq:n_u_minco_b}).

For the normalized $\bar{p}(t)$, the mapping is written by Eq.(\ref{eq:n_u_minco_mapping}), where the banded matrix $\mathbf{M}(\mathbf{1})$ becomes constant because the quantity of the pieces $M$ is fixed during trajectory optimization.
Therefore, $\mathbf{M}(\mathbf{1})^{-1}$ can be computed explicitly or performed PLU factorization offline, instead of online factorization every time $T$ changes in the optimization.
Then the coefficients $\bar{\mathbf{C}}$ can be obtained from Eq.(\ref{eq:n_u_minco_c}) and the coefficients $\hat{\mathbf{C}}$ of the original trajectory can be obtained by scaling the normalized trajectory, as written in Eq.(\ref{eq:u_minco_c}).
Moreover, the gradient calculation mentioned in Sec.~\ref{subsec:minco} can be performed without online factorization either.
These have an improvement in computational efficiency, as shown in Sec~.\ref{sec:Evaluations:Ablation:minco}.

To summarize, to use time-uniform MINCO yet avoid online PLU factorization, we first normalize the trajectory in time. 
Then we use the property that $\mathbf{M}(\mathbf{1})$ of the normalized trajectory can be processed offline to quickly obtain the normalized parameters $\bar{\mathbf{C}}$ and gradients. 
Finally, we get the parameters $\hat{\mathbf{C}}$ and gradients of the time-uniform MINCO which is truly desired by time deflating as shown in Eq.(\ref{eq:u_minco_c}).

\section{Polynomial-based Online Planning}
\label{sec:Online_Replan}


In this section, we use time-uniform MINCO as the trajectory representation for online planning, requiring the shortest possible execution time while satisfying some environmental and actuator constraints.
First in Sec.~\ref{sec:Problem_Formulation} we construct an optimization problem considering above requirements.
Then based on our dealing with its inequality and equation constraints respectively in Sec.~\ref{sec:Inequality_Constraints_Transcription} and \ref{sec:Equality_Constrains_Elimination}, the problem is reformulated into an unconstrained optimization problem in Sec.~\ref{sec:Problem_Reformulation}.
Additionally, in Sec.~\ref{subsec:Implement_Details}, we state some engineering considerations that are effective in improving racing performance.
Note that the symbol definitions of time-uniform MINCO in this section are inherited from Sec.~\ref{sec:traj}.

\subsection{Problem Formulation}
\label{sec:Problem_Formulation}

\begin{figure}[!t]
	\centering
    \vspace{0.2cm}
	\includegraphics[width=1\linewidth]{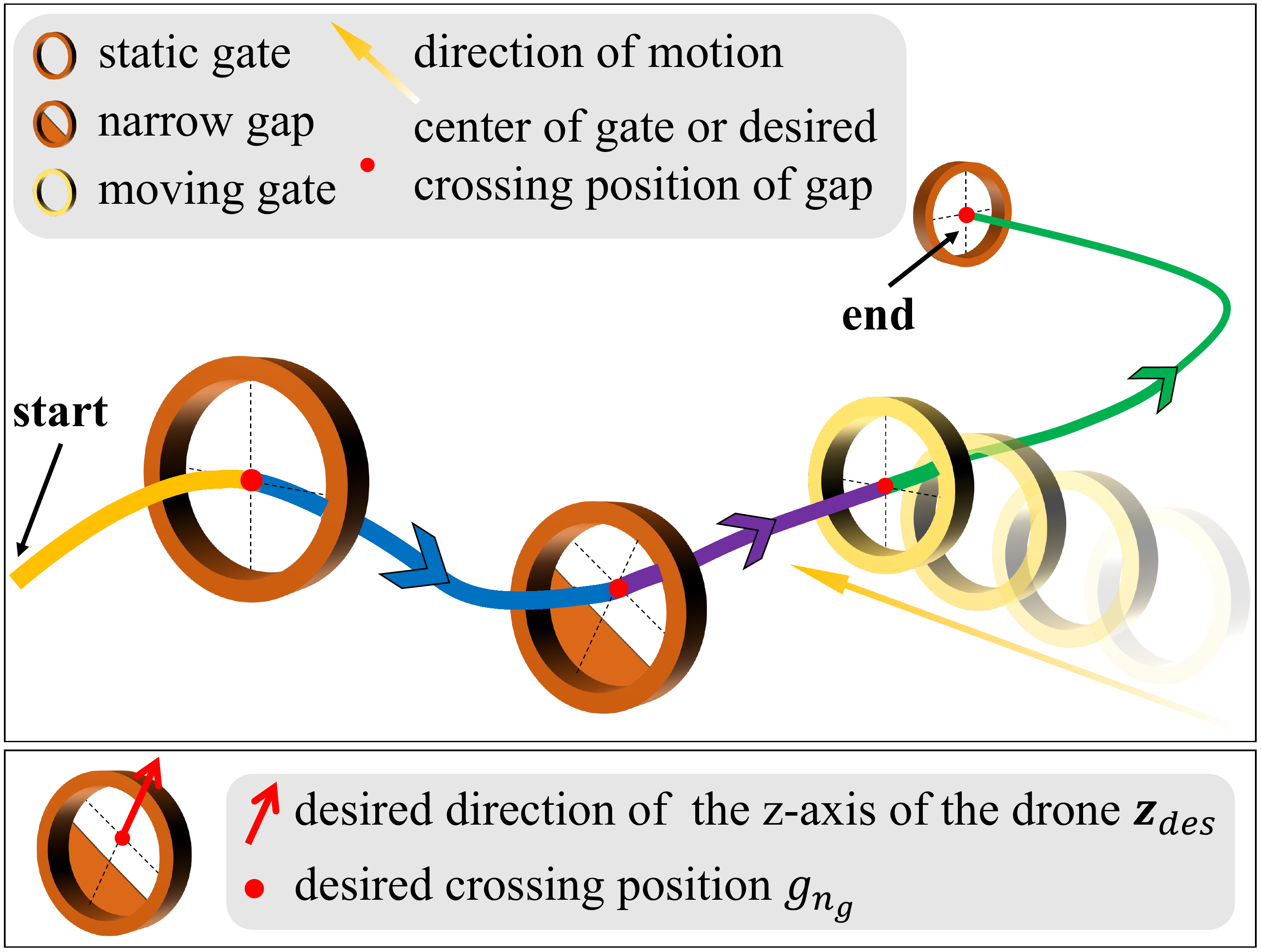}
    \captionsetup{font={footnotesize}}
	\caption{
        Illustration of the segmented trajectory for replan to traversal the four gates in order.
	}
	\label{fig:traj}
    \vspace{-1cm}
\end{figure}

In this paper, we use segmented polynomials to represent the replan trajectory.
Given the next $N$ gates, we plan an $N$-segment trajectory.
As shown in Fig.~\ref{fig:traj}, each colored curve represents one segment.
The $n$-th segment $\sigma_n(t)$ is an $s$-order $m$-dimensional $M_n$-piece time-uniform MINCO, 
whose coefficients and intermediate points are defined as $\hat{\mathbf{C}}_n$ and $\mathbf{Q}_n$ respectively, detailed in Eq.(\ref{eq:seg_parameter}).
Its time allocation is defined as $\hat{\mathbf{T}}_n=(T_n/M_n)\mathbf{1}$, where $T_n$ is the trajectory duration of $n$-th segment.
The whole segmented trajectory $\sigma(t):[t_0, t_N]\mapsto\mathbb{R}^m$ is formulated as
\begin{equation}
    \begin{gathered}
        \sigma(t_{n-1}+t) = \sigma_n(t), \\
        \forall n \in \{1,2,\dots,N\}, ~\forall t \in [0,T_n],
    \end{gathered}
\end{equation}
where $t_n=t_0+\sum_{j=1}^{n}T_j$ is the timestamp and $t_0$ is the start time of the trajectory.
The coefficients, time allocation, and intermediate points of the whole trajectory can be written as
\begin{equation}
    \begin{aligned}
        \boldsymbol{\mathcal{C}} &= (\hat{\mathbf{C}}_1\tp,\dots,\hat{\mathbf{C}}_n\tp,\dots,\hat{\mathbf{C}}_{N}\tp)\tp &&\in \mathbb{R}^{(\sum_{n=1}^{N}2 M_n s) \times m},\\
        \boldsymbol{\mathcal{T}} &= (\hat{\mathbf{T}}_1\tp,\dots,\hat{\mathbf{T}}_n\tp,\dots,\hat{\mathbf{T}}_{N}\tp)\tp &&\in \mathbb{R}_{>0}^{(\sum_{n=1}^{N}M_n)},\\
        \boldsymbol{\mathcal{Q}} &= (\mathbf{Q}_1,\dots,\mathbf{Q}_n,\dots,\mathbf{Q}_{N}) &&\in \mathbb{R}^{m \times \left(\sum_{n=1}^{N}(M_n-1)\right)}.
    \end{aligned}
\end{equation}

We require the trajectory to pass through a series of gates, both static and moving, as quickly as possible, with constraints of dynamical feasibility, narrow gap crossing, and dynamic obstacle avoidance.
Taking all requirements into account, our problem takes the following form:
\begin{subequations}
    \label{eq:problem_formulation}
    \begin{align}
        \underset{\boldsymbol{\mathcal{C}},\boldsymbol{\mathcal{T}}}{min}~~
                 & \label{eq:problem_formulation:cost} J_o=\int_{t_0}^{t_N} \| \sigma^{(s)} (t) \|^2 dt + \rho \cdot ||\boldsymbol{\mathcal{T}}||_1,                                 \\
        s.t.~~
                & \label{eq:problem_formulation:boundary_1} \sigma_1^{[0,s-1]}(0)=\mathbf{s}^o,                                                                       \\
                 & \label{eq:problem_formulation:boundary_2} \sigma_N^{[0,s-1]}(T_N)=\mathbf{s}^f,                                                                     \\
                 & \label{eq:problem_formulation:boundary_3} \sigma_{n}^{[0,s-1]}(T_{n})=\sigma_{n+1}^{[0,s-1]}(0),~\forall n \in \{1,\dots,N-1\},                         \\
                 & \label{eq:problem_formulation:gate} \sigma(t_{n})=g_n(t_{n}), ~\forall n \in \{1,\dots,N\},                                             \\
                 & \label{eq:problem_formulation:inequal} \mathcal{G}_x\left(\sigma(t),...,\sigma^{(s)}(t),t\right)\preceq \mathbf{0}, ~\forall x \in \mathcal{X},\forall t \in [t_0,t_N],
    \end{align}
\end{subequations}
where we define two costs in Eq.(\ref{eq:problem_formulation:cost}) for smoothness and short execution time, which are weighed by parameter $\rho$.
Eq.(\ref{eq:problem_formulation:boundary_1}--\ref{eq:problem_formulation:boundary_2}) and Eq.(\ref{eq:problem_formulation:boundary_3}) are the boundary conditions and the continuity constraint up to degree $s-1$.
$\mathbf{s}^o$ and $\mathbf{s}^f$ are boundary states. 
We denote $\sigma^{[x,y]} \in \mathbb{R}^{m \times \left(y-x+1\right)}$ as
\begin{equation}
    \sigma^{[x,y]}=\left(\sigma^{\left(x\right)},\sigma^{\left(x+1\right)},\dots,\sigma^{\left(y\right)} \right),~~x < y.
\end{equation}
Moreover, Eq.(\ref{eq:problem_formulation:gate}) is the gate-through constraint, where $g_i(t)$ is the predicted trajectory for the $i$-th gate.
Eq.(\ref{eq:problem_formulation:inequal}) is continuous-time constraints, the set $\mathcal{X}=\{t,b,g,d\}$ include actuator physical limits on thrust $(t)$ and body rate $(b)$, narrow gap crossing $(g)$ and dynamic obstacle avoidance~$(d)$.

\subsection{Inequality Constraints Transcription}
\label{sec:Inequality_Constraints_Transcription}
For the inequality constraints Eq.(\ref{eq:problem_formulation:inequal}), which are required to be fulfilled along the whole trajectory.
We use the thought behind penalty function method~\cite{jennings1990computational} to deal with these constraints becoming time integral of constraint violations, which is then evaluated by a finite sum of sample points.
We define these points sampled on $n$-th segment by $\mathring{\mathbf
p}_{n,j}=\sigma_n((j/\kappa_n)T_n), j\in \{0,1,2,\dots,\kappa_n\}$, where $\kappa_n$ is the sample quantity and we name $\mathring{\mathbf
p}_{n,j}$ as \textbf{constraint points}.
We denote the penalty function of the constraint points as
\begin{equation}
    \mathcal{G}_x({n,j})=\mathcal{G}_x\left(\mathring{\mathbf
    p}_{n,j},\mathring{\mathbf p}_{n,j}^{(1)},...,\mathring{\mathbf p}_{n,j}^{(s)},t_{n-1}+\frac{j}{\kappa_n}T_n\right).
\end{equation}
Then these inequality constraints Eq.(\ref{eq:problem_formulation:inequal}) can be transformed into a weighted sum of the sampled penalty as
\begin{equation}
\begin{gathered}
    \label{eq:inequality_cost}
    \underset{\boldsymbol{\mathcal{C}},\boldsymbol{\mathcal{T}}}{min}~~\sum\nolimits_{x}^{\mathcal{X}}\lambda_x J_x,\\
    J_x=\sum\nolimits_{n=1}^{N}\frac{T_n}{\kappa_n}
    \sum\nolimits_{j=0}^{\kappa_n}\bar\omega_j\underset{}{max}
    (\mathcal{G}_x(n,j),0)^3,
\end{gathered}
\end{equation}
where $\lambda_x$ is the weight for each cost $J_x$, we follow the trapezoidal rule~\cite{press2007numerical} $(\bar{\omega}_0,\bar{\omega}_1,\dots,\bar{\omega}_{\kappa_i})=(1/2,1,\cdots,1,1/2)$.
Then for the inequality constraints Eq.(\ref{eq:problem_formulation:inequal}), with Eq.(\ref{eq:minco_def},~\ref{eq:inequality_cost}), 
once the gradients of $\mathcal{G}_x(n,j)$ w.r.t. $\mathring{\mathbf p}_{n,j}^{(k)}, k \in \{1,\dots,s\}$ and $\{T_1,\dots,T_n\}$ are given, we can derive the gradients $\partial J_x/ \partial \boldsymbol{\mathcal{C}}$ and $\partial J_x/ \partial \boldsymbol{\mathcal{T}}$ which are required in the optimization.
Then we introduce each penalty function and its gradient.

\subsubsection{Actuator Limits with Drag Effects $\mathcal{G}_t$ and $\mathcal{G}_b$}
\label{sec:limits}
To ensure that the trajectory is physically feasible, we constrain the thrust~$f$ and body rate $\omega$ through the differential flatness.
Meanwhile, due to the high speed of racing, aerodynamic effects cannot be ignored, then we extended the quadrotor's dynamics with a drag model of our previous work~\cite{wang2022robust}.
As shown in the Eq.(17-21) of \cite{wang2022robust}, given the trajectory, thrust, body rate and rotation $R \in SO(3)$ in the world frame of the constraint point $\mathring{\mathbf p}_{n,j}$ can be denoted as
\begin{equation}
\label{eq:flatness}
    f,\omega,R = \mathcal{F}\left(\mathring{\mathbf p}_{n,j}^{(1)}, \mathring{\mathbf p}_{n,j}^{(2)},\mathring{\mathbf p}_{n,j}^{(3)}\right),
\end{equation}

We define the penalty of actuator physical limits as
\begin{equation}
\label{eq:physical_limits}
\begin{aligned}
    \mathcal{G}_t(n,j)&=
     (f-f_m)^2-f_r^2,\\
    \mathcal{G}_b(n,j) &= ||\omega||^2-\omega_{max}^2,
\end{aligned}
\end{equation}
where $f_m=(f_{max}+f_{min})/2$ and $f_r=(f_{max}-f_{min})/2$, $f_{max}$ and $f_{min}$ are the maximum and minimum values of thrust.
$\omega_{max}$ is the maximum body rate. With Eq.(\ref{eq:flatness},~\ref{eq:physical_limits}), the gradients $\partial \mathcal{G}_t(n,j)/\partial \mathring{\mathbf p}_{n,j}^{(k)},~ \partial \mathcal{G}_b(n,j)/\partial \mathring{\mathbf p}_{n,j}^{(k)},~k={1,2,3}$ can be calculated easily using the chain rule.



\subsubsection{Narrow Gap Crossing $\mathcal{G}_g$}
\label{sec:Narrow_Gap}
In this framework, we assume that we can obtain the optimal attitude and position for crossing a narrow gap from other modules such as online detection.
As shown in Fig.~\ref{fig:traj}, we set the desired direction of the z-axis of the quadrotor when traversing the gate as a normalized vector $\mathbf{z}_{des} \in \mathbb{R}^m$. 
To avoid loss of generality, we set the optimal crossing position as a gate, denoted as the $n_g$-th gate to be passed through.
The position constraint will be detailed in Sec.~\ref{sec:gate_con}, and here we only describe the direction constraint during gate crossing.
To make it safer to traverse the narrow gate with a large attitude, we impose this constraint on the trajectory at a certain sample range $n_{ran}$ in front of and behind the $n_g$-th gate:
\begin{equation}
\label{eq:narrow_gap}
\begin{gathered}
    \mathcal{G}_g(n,j)=|| R\mathbf{e}_3 - \mathbf{z}_{des} ||^2-\theta_{tol},\\
    \forall j \in 
    \left\{
    \begin{aligned}
        &\{0,1,\dots,n_{ran}\}, &&n=n_g\\
        &\{\kappa_n-n_{ran},\dots,\kappa_n\}, &&n=n_g-1
    \end{aligned}
    \right.,
\end{gathered}
\end{equation}
where $\mathbf{e}_3=(0,0,1)\tp$, $\theta_{tol} \in (0,1)$ is the tolerance, $R$ is the rotation of $\mathring{\mathbf p}_{n,j}$ obtained from Eq.(\ref{eq:flatness}).
Then the gradien $\partial \mathcal{G}_g(n,j)/\partial \mathring{\mathbf p}_{n,j}^{(k)}$ can be computed with Eq.(\ref{eq:flatness},~\ref{eq:narrow_gap}).

\subsubsection{Dynamic Obstacle Avoidance $\mathcal{G}_d$}
Based on our previous work~\cite{wang2021autonomous}, we model a dynamic obstacle as ellipsoid:
\begin{equation}
    \mathbb{E}=\{\mathbf{x}|E(\mathbf{x},\mathbf{y})<1\},~~
    E(\mathbf{x},\mathbf{y})=(\mathbf{x}-\mathbf{y})\tp\mathbf{H}(\mathbf{x}-\mathbf{y}),
\end{equation}
where $\mathbf{H}=R_E\tp diag(1/a^2,1/b^2,1/c^2)R_E$, $a,b,c$ is the axis-length, $R_E$ is its rotation and $\mathbf{y}$ is the center of the ellipsoid.
Then we define the obstacle avoidance penalty as
\begin{equation}
    \label{eq:dynamic}
    \mathcal{G}_d(n,j)=d_{thr}-E\left(\mathring{\mathbf p}_{n,j},~p_d(t_{n-1}+\frac{j}{\kappa_n}T_n)\right),
\end{equation}
where $p_d(t)$ is the predicted trajectory of the dynamic obstacle, $d_{thr}$ is security threshold.
Similar to Sec.~\ref{sec:limits} and \ref{sec:Narrow_Gap}, by the chain rule, we can utilize Eq.(\ref{eq:dynamic}) to obtain gradient of $\mathcal{G}_d(n,j)$ w.r.t. $\mathring{\mathbf p}_{n,j}$ and $\{T_1,\dots,T_n\}$, which are used to derive $\partial J_d/ \partial \boldsymbol{\mathcal{C}}$ and $\partial J_d/ \partial \boldsymbol{\mathcal{T}}$ for optimization, as stated in Sec.~\ref{sec:Inequality_Constraints_Transcription}.

\subsection{Equality constraints Elimination}
\label{sec:Equality_Constrains_Elimination}
In Sec.~\ref{sec:Inequality_Constraints_Transcription} we transform the inequality constraints by the penalty function method, in this section we eliminate the equation constraints by replacing the decision variables.
\subsubsection{Boundary Conditions and Continuity Constraint}
\label{sec:Boundary_Conditions}
We define $\mathbf{z}_0$ as the start state of the first segment and $\mathbf{z}_n $ as the end state of the $n$-th segment:
\begin{equation}
    \begin{aligned}
    \mathbf{z}_0&=\sigma_{1}^{[0,s-1]}(0),\\
    \mathbf{z}_n&=\sigma_{n}^{[0,s-1]}(T_{n}),~\forall n \in \{1,\dots,N\}.
    \end{aligned}
\end{equation}
For the $n$-th segment, given the time allocation $\hat{\mathbf{T}}_n$, intermediate points $\mathbf{Q}_n$ and the boundary conditions $\mathbf{z}_{n-1}$,~$\mathbf{z}_n$, its coefficients $\hat{\mathbf{C}}_n$ can be uniquely determined by a serie of calculations Eq.(\ref{eq:u_minco_mapping}--\ref{eq:n_u_minco_c}).
We denote this calculation process as
\begin{equation}
\label{eq:mapping}
        \mathbf{C}_n =\mathcal{M}(\mathbf{Q}_n,\hat{\mathbf{T}}_n,\mathbf{z}_{n},\mathbf{z}_{n-1}).
\end{equation}

We set $\boldsymbol{\mathcal{Z}} = (\mathbf{z}_1, \dots, \mathbf{z}_{N-1})$, and fix $\mathbf{z}_0=\mathbf{s}^o$ and $\mathbf{z}_N=\mathbf{s}^f$.
Based on Eq.(\ref{eq:mapping}), $\boldsymbol{\mathcal{C}}$ can be determined with variables $\boldsymbol{\mathcal{Z}}$, $\boldsymbol{\mathcal{T}}$ and $\boldsymbol{\mathcal{Q}}$.
Therefore, for our original problem Eq.(\ref{eq:problem_formulation}), we replace the decision variables from $(\boldsymbol{\mathcal{C}}$, $\boldsymbol{\mathcal{T}})$ to $(\boldsymbol{\mathcal{Z}}$, $\boldsymbol{\mathcal{T}}$, $\boldsymbol{\mathcal{Q}})$, eliminating the constraints of Eq.(\ref{eq:problem_formulation:boundary_1}--\ref{eq:problem_formulation:boundary_2}).

\subsubsection{Gate-through Constraint}
\label{sec:gate_con}


The requirement of crossing gate is usually handled with soft constraints, for instance, penalty function $\|\sigma(t_{n})-g_n(t_{n})\|^2$ used in \cite{romero2022model,romero2022time}.
However, As shown in Fig.~\ref{fig:traj}, when the gate is small, the space tolerance at the moment of crossing is very low.
The use of soft constraints does not guarantee that trajectory passes through the center of the gate, which leads to more pressure on safety being placed on other modules, such as the controller.
In this paper, we construct hard constraints for gate-through to ensure that the trajectory can traverse the center of the gate.

We split the state matrix $\mathbf{z}_n$ as $\left((\mathbf{z}_n)_{0}, (\mathbf{z}_n)_{*}\right)$, where $(\mathbf{z}_n)_{0} \in \mathbb{R}^{m \times 1}$ is the $0$-order derivative state, and $(\mathbf{z}_n)_{*} \in \mathbb{R}^{m \times (s-1)}$ means the derivatives whose order from $1$ to $s$.
Then the gate-through constraints Eq.(\ref{eq:problem_formulation:gate}) can be written as
\begin{equation}
    \begin{aligned}
        (\mathbf{z}_n)_{0}
            &=g_n(t_{n})\\
            &=g_n(t_0+\sum\nolimits_{j=1}^{n}T_j),
    \end{aligned}
\end{equation}
which demonstrates that $(\mathbf{z}_n)_{0}$ can determined by $\boldsymbol{\mathcal{T}}$.
Therefore, the mapping function Eq.(\ref{eq:mapping}) can be formulated as
\begin{equation}
    \begin{aligned}
        \label{eq:new_mapping}
        \mathbf{C}_n=\mathcal{M}\Bigg(\mathbf{Q}_n,\hat{\mathbf{T}}_n,
        \Big(g_{n}(t_0+\sum\nolimits_{j=1}^{n}T_j),~(\mathbf{z}_{n})_{*}&\Big),\\
        \Big(g_{n-1}(t_0+\sum\nolimits_{j=1}^{n-1}T_j),~(\mathbf{z}_{n-1})_{*}&\Big)\Bigg).
    \end{aligned}
\end{equation}
To eliminate the constraints of Eq.(\ref{eq:problem_formulation:gate}), We replace the variables in Sec.~\ref{sec:Boundary_Conditions} from $(\boldsymbol{\mathcal{Z}}$, $\boldsymbol{\mathcal{T}}$, $\boldsymbol{\mathcal{Q}})$ to $(\boldsymbol{\mathcal{Z}}_*$,~$\boldsymbol{\mathcal{T}}$,~$\boldsymbol{\mathcal{Q}})$, where ${\boldsymbol{\mathcal{Z}}_*}=((\mathbf{z}_1)_{*},\dots, (\mathbf{z}_{N-1})_{*})$.

\subsection{Problem Reformulation}
\label{sec:Problem_Reformulation}
Combining the transcription of inequality constraints in Sec.~\ref{sec:Inequality_Constraints_Transcription} and the elimination of equality constraints in Sec.~\ref{sec:Equality_Constrains_Elimination}, the whole original problem Eq.(\ref{eq:problem_formulation}) is finally reformulated as an unconstrained optimization problem:
\begin{equation}
    \label{eq:unconstrained_optimization_problem}
    \underset{\boldsymbol{\mathcal{Z}}_*,\boldsymbol{\mathcal{T}},\boldsymbol{\mathcal{Q}}}{min}
    J=(J_o+
    \sum\nolimits_{x}^{\mathcal{X}}\lambda_x J_x ).
\end{equation}

Sec.~\ref{sec:Inequality_Constraints_Transcription} gives the gradients of $J$ w.r.t. $(\boldsymbol{\mathcal{C}},\boldsymbol{\mathcal{T}})$,  but we change the decision variables in Eq.(\ref{eq:unconstrained_optimization_problem}).
Based on Eq.(\ref{eq:mapping}), we can obtain the gradients of $\boldsymbol{\mathcal{C}}$ w.r.t. $(\boldsymbol{\mathcal{Z}}$,~$\boldsymbol{\mathcal{T}}$,~$\boldsymbol{\mathcal{Q}})$, which are detailed in the Eq.(60,~68) in \cite{wang2022geometrically}.
However, in the new mapping function Eq.(\ref{eq:new_mapping}), we set $\mathbf{z}_n$ a matrix with $(\mathbf{z}_n)_{0}$ and $(\mathbf{z}_n)_*$ as variables, and $(\mathbf{z}_n)_{0}$ is a vector with $\boldsymbol{\mathcal{T}}$ as decision variable.
Therefore, to obtain the gradient of $J$ w.r.t. the new decision variables $(\boldsymbol{\mathcal{Z}}_*$,~$\boldsymbol{\mathcal{T}}$,~$\boldsymbol{\mathcal{Q}})$, which are required when solving the final problem Eq.(\ref{eq:unconstrained_optimization_problem}), we derive the gradient 
\begin{equation}
    \begin{aligned}
        \frac{\partial{(\mathbf{z}_n)_0}}{\partial T_k}=\frac{\partial g_{n}(t_n)}{\partial t_n}\frac{\partial (t_0+\sum\nolimits_{j=1}^{n}T_j)}{\partial T_k}=\left\{
        \begin{aligned}
            &0,&n < k\\
            &\dot{g}_{n}(t_n),&n \geq k
        \end{aligned}
        \right. .
    \end{aligned}
\end{equation}


\subsection{Implementation Details}
\label{subsec:Implement_Details}
\subsubsection{Parallel Optimization for Different Topologies}
Although dynamic obstacles are addressed in many works~\cite{wang2021autonomous,chen2022rast}, they only care about obstacle avoidance, ignoring the spatial topology segmentation brought by dynamic obstacles.
The choice of different topologies has a significant impact on the execution time of racing, as demonstrated in Sec.~\ref{sec:Evaluations:Ablation:topo}.
Therefore, in this paper, as shown in Fig.~\ref{fig:xiaorong3}, we generate several trajectories with different topologies split by the dynamic obstacle.
Then we use them in parallel as initial values for trajectory optimization and finally choose the one with the shortest flight time for the drone to execute.

\subsubsection{Numerical Optimization}
We adopt L-BFGS\footnote{https://github.com/ZJU-FAST-Lab/LBFGS-Lite}~\cite{liu1989limited} to solve the unconstrained optimization problem Eq.(\ref{eq:unconstrained_optimization_problem}).

\subsubsection{Global Planning}
For some large-scale scenarios, such as Sec.~\ref{sec:Evaluations:dji}, we first generate a global reference trajectory using a one-segment MINCO trajectory, and then use part of the global trajectory as the initial value for the optimization of the local online replanning during flight.

\section{Evaluations}
\label{sec:Evaluations}
In this section, to verify the effectiveness of our contributions summarized in Sec.~\ref{sec:Introduction}, we design three ablation experiments.
Then to validate the practicality of our planning method, we integrate the method into a complete quadrotor system, which is applied to 2022 DJI Robomaster Intelligent UAV Championship\textsuperscript{\ref{dji_bisai_web}}.
All the simulation experiments are run on a desktop equipped with
an Intel Core  i7-10700 CPU.

\subsection{Ablation Experiments}
\label{sec:Evaluations:Ablation}

\subsubsection{Evaluation for Efficiency Improvement of Time-uniform MINCO}
\label{sec:Evaluations:Ablation:minco}

\begin{figure}[!t]
	\centering
    \vspace{0.0cm}
	\includegraphics[width=1\linewidth]{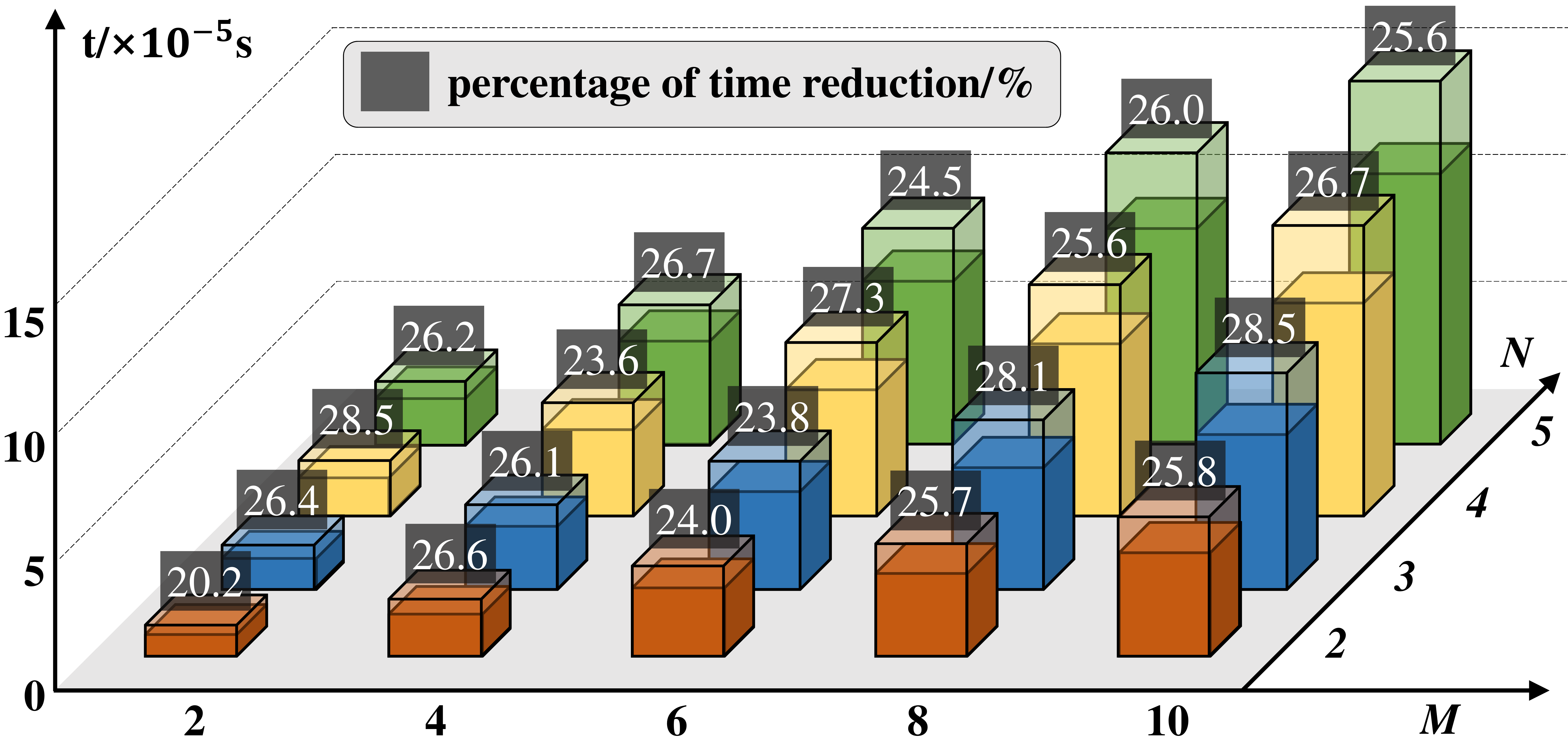}
    \captionsetup{font={footnotesize}}
	\caption{
        Illustration of the time of a single iteration in the trajectory optimization.
        The transparent area represents the more computation time that MINCO takes than the time-uniform MINCO.
	}
	\label{fig:xiaorong1}
    \vspace{-1.4cm}
\end{figure}

We compare the computational efficiency of MINCO and time-uniform MINCO with the same segment and piece number.
For both of them, we test the number of segments $N$ varied from $2$ to $5$, while the number of pieces $M=\{2,4,\dots,10\}$.
Both methods are given the same initial trajectory for each comparison.
The results are illustrated in Fig.~\ref{fig:xiaorong1}, where the time of a single iteration of the trajectory optimization is visualized.
We represent the time spent by time-uniform MINCO with different colored bars.
The transparent part indicates the additional time required for MINCO compared to time-uniform MINCO.
Additionally, we show the percentage of time reduction of time-uniform MINCO over MINCO on each bar.

As shown in Fig.~\ref{fig:xiaorong1}, using time-uniform MINCO effectively improves the calculation efficiency, even if the quantity of segments or pieces varies.

\begin{figure}[htbp]
	\centering
	\includegraphics[width=1\linewidth]{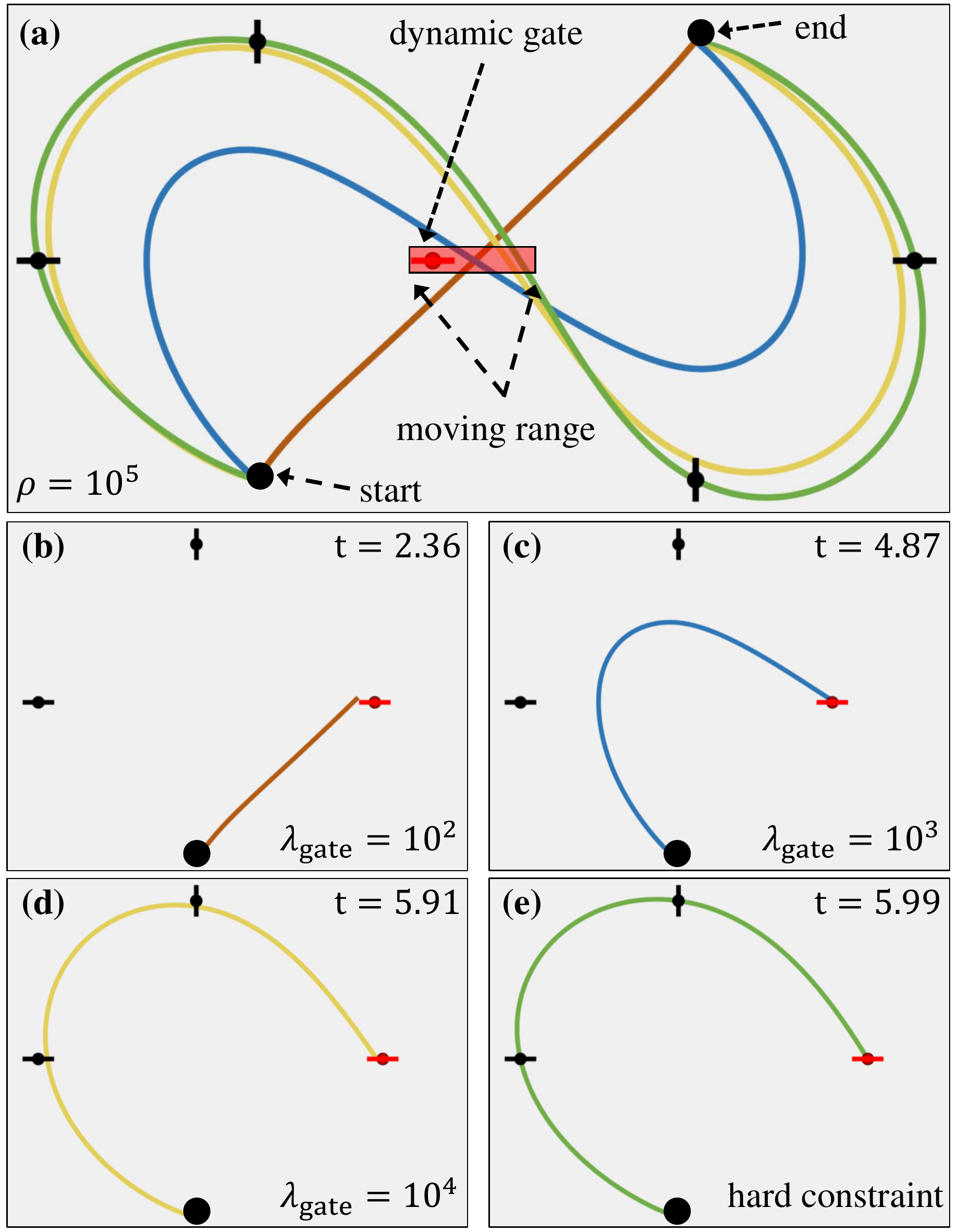}
    \captionsetup{font={footnotesize}}
	\caption{
        Illustration of the top view of the results in Sec.~\ref{sec:Evaluations:Ablation:hard}.
        The short thick black lines represent the static gates and the short thick red line represents the dynamic gate.
        The circle on the line represents the center of the gate.
        The red bar on the red line indicates the movement range of the dynamic gate.
        \textbf{(a)} demonstrates the trajectories of the quadrotor completing the track under different soft constraint weights and the same time weight $\rho=10^5$.
        \textbf{(b)}-\textbf{(e)} represent the moments for the quadrotor to pass through the dynamic gate with different parameters.
	}
	\label{fig:xiaorong2_traj}
\end{figure}

\subsubsection{Evaluation for constraints for Precise Gate Crossing}
\label{sec:Evaluations:Ablation:hard}
We compare our method proposed in Sec.~\ref{sec:gate_con} with a commonly used soft constraint method which uses penalty function $\lambda_{gate}\|\sigma(t_{n})-g_n(t_{n})\|^2$, where $\lambda_{gate}$ is the weight.
The experiment scenario is set up as shown in Fig.~\ref{fig:xiaorong2_traj}.\textbf{(a)}, where the quadrotor is required to pass through $5$ gates in order.
The red dynamic gate moves at $2$ m/s horizontally back and forth from side to side within a certain range, which is indicated by the red bar in Fig.~\ref{fig:xiaorong2_traj}.\textbf{(a)}.
To compare fairly, each trajectory optimization is given the same initial value and we set other parameters that are not about this experiment the same.
We set different time weight $\rho$ and soft constraint weights $\lambda_{gate}$ for the experiments.

\begin{figure}[tp]
	\centering
	\includegraphics[width=1\linewidth]{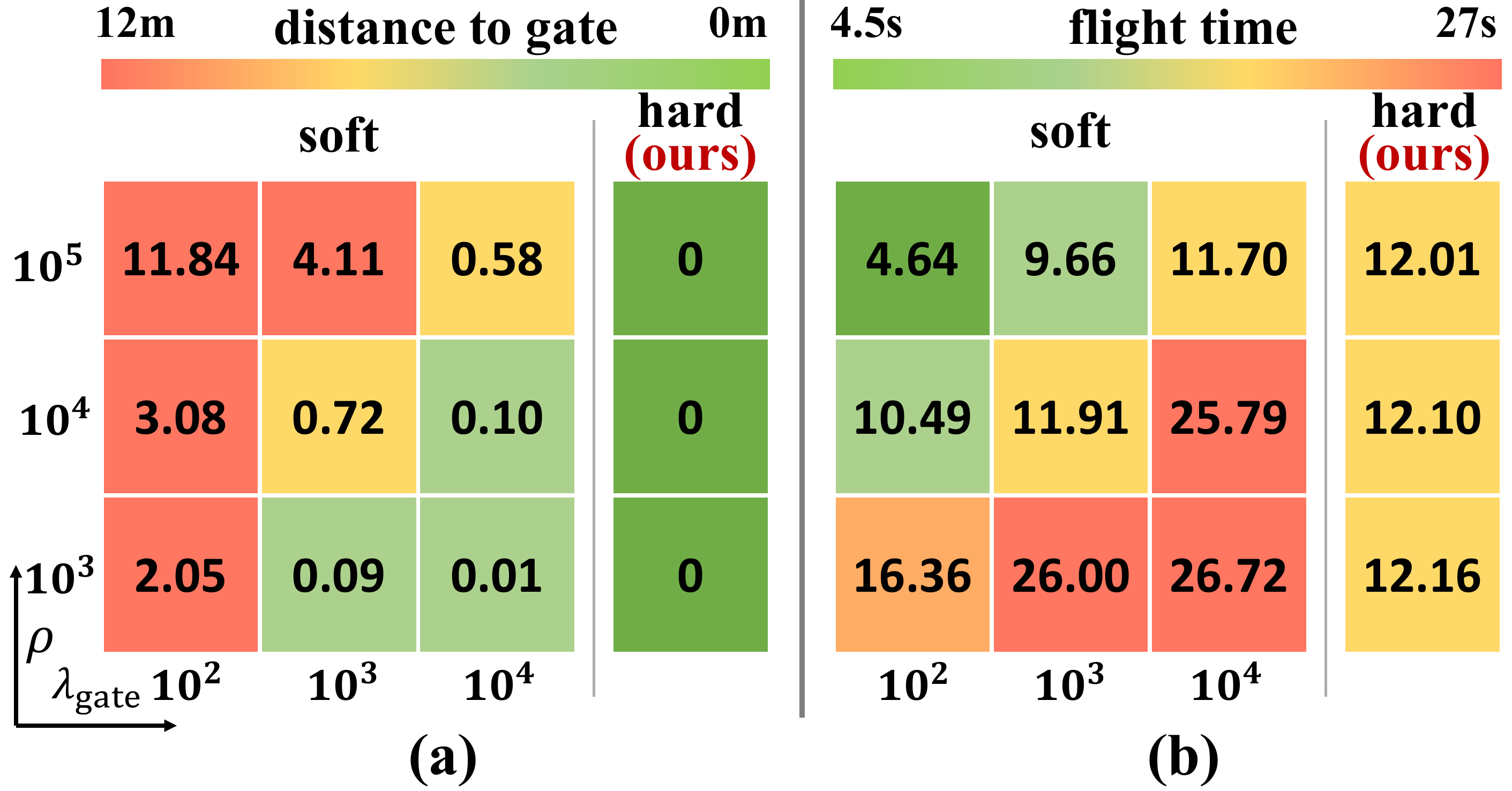}
    \captionsetup{font={footnotesize}}
	\caption{
        Results of experiments in Sec.~\ref{sec:Evaluations:Ablation:hard}, where time weight $\rho$ and soft constraint weight $\lambda_{gate}$ vary from $10^3$ to $10^5$ and $10^2$ to $10^4$, respectively.
        \textbf{(a)} shows the average distance from the trajectory to the corresponding $n$-th gate at the moment $t_n$, from far to near, the color changes from red to green.
        \textbf{(b)} displays the flight time with different parameters, from less to more, and the color varies from green to red.
	}
	\label{fig:xiaorong2}
    \vspace{-1cm}
\end{figure}

The results are illustrated in Fig.~\ref{fig:xiaorong2}, soft constraint method struggles to find suitable parameters to trade off the shortest possible execution time with precise gate crossing.
To elaborate, when choosing the parameters that achieve a small distance to gate as shown in the green area of Fig.~\ref{fig:xiaorong2}.\textbf{(a)}, the soft constraint method causes poor results in terms of the time of flight as shown in the red area of Fig.~\ref{fig:xiaorong2}.\textbf{(b)}.
Conversely, when choosing the parameters that aim for the shortest possible flight time in the green or yellow area of Fig.~\ref{fig:xiaorong2}.\textbf{(b)}, the soft constraint method is difficult to precisely crossing the gate, which corresponds to the red area of Fig.~\ref{fig:xiaorong2}.\textbf{(a)}.
In contrast, our hard constraint approach only needs to focus on adjusting the time weight $\rho$ to shorten the execution time, while maintaining precise traversal.

To more visually demonstrate the importance of accurate traversal, in Fig.~\ref{fig:xiaorong2_traj}, we visualize the trajectories of our method and the soft constraint method when the time weight $\rho=10^5$.
The results show that, compared to the soft constraint method, our method guarantees accurate traversal for both dynamic and static gates.

\subsubsection{Evaluation for Choosing Different Topologies Segmented by Dynamic Obstacles}
\label{sec:Evaluations:Ablation:topo}
\begin{figure}[!t]
	\centering
    \vspace{0.0cm}
	\includegraphics[width=1\linewidth]{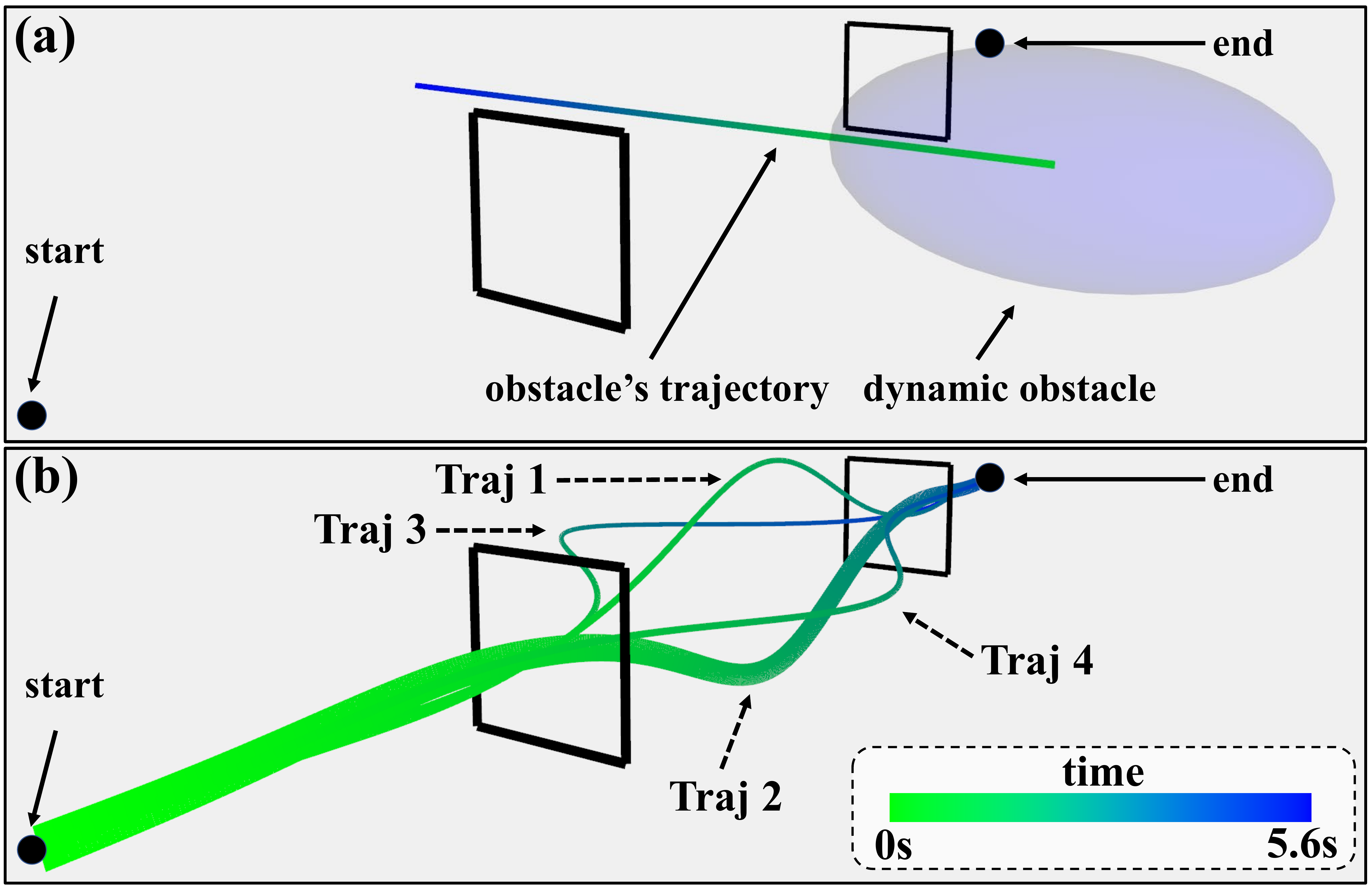}
    \captionsetup{font={footnotesize}}
	\caption{
        Illustration of the results in Sec.~\ref{sec:Evaluations:Ablation:topo}.
        \textbf{(a)} demonstrates the experiment setup.
        \textbf{(b)} shows the trajectories optimized based on initial values from different topological.
	}
	\label{fig:xiaorong3}
\end{figure}

We use a typical scenario to verify the usefulness of this engineering consideration.
As shown in Fig.~\ref{fig:xiaorong3}.\textbf{(a)}, the quadrotor is required to take off from the start point, pass through two gates in order, and then reach the end point.
Between the two gates, there is a dynamic obstacle which we model as an ellipsoid. 
The movement of its central position with respect to time we represent by a colored trajectory in Fig.~\ref{fig:xiaorong3}.\textbf{(a)}.
We set the trajectories through different topologies past the obstacle as initial values.
The final trajectories after optimization are illustrated in Fig.~\ref{fig:xiaorong3}.\textbf{(b)},  the bolded trajectory is significantly faster than the trajectories of other topologies.
Meanwhile, we present the execution time of each trajectory in Tab.~\ref{tab:topo}, where the choice of different topologies has a considerable impact on the execution time, which confirms the validity of our evaluating and choosing different topologies segmented by the dynamic obstacle.

\begin{table}[!t]
	\vspace{-0.2cm}
	\centering
	\caption{Trajectories of Different Topologies} 
    \tabcolsep=0.35cm 
	\begin{tabular}{c|cccc}
		\toprule
		Traj number & 1 & 2 & 3 & 4  \\
		\midrule
		execution time  (s) & 3.82 & \bf 3.79 & 5.60 & 4.38   \\
		trajectory length (m) &  20.41 & \bf 20.05 & 22.97 & 20.64  \\
		\bottomrule
	\end{tabular}
	\label{tab:topo}
	\vspace{-1.2cm}
\end{table} 

\subsection{2022 DJI Robomaster Intelligent UAV Championship}
\label{sec:Evaluations:dji}
\begin{figure}[!t]
	\centering
    \vspace{0.2cm}
	\includegraphics[width=1\linewidth]{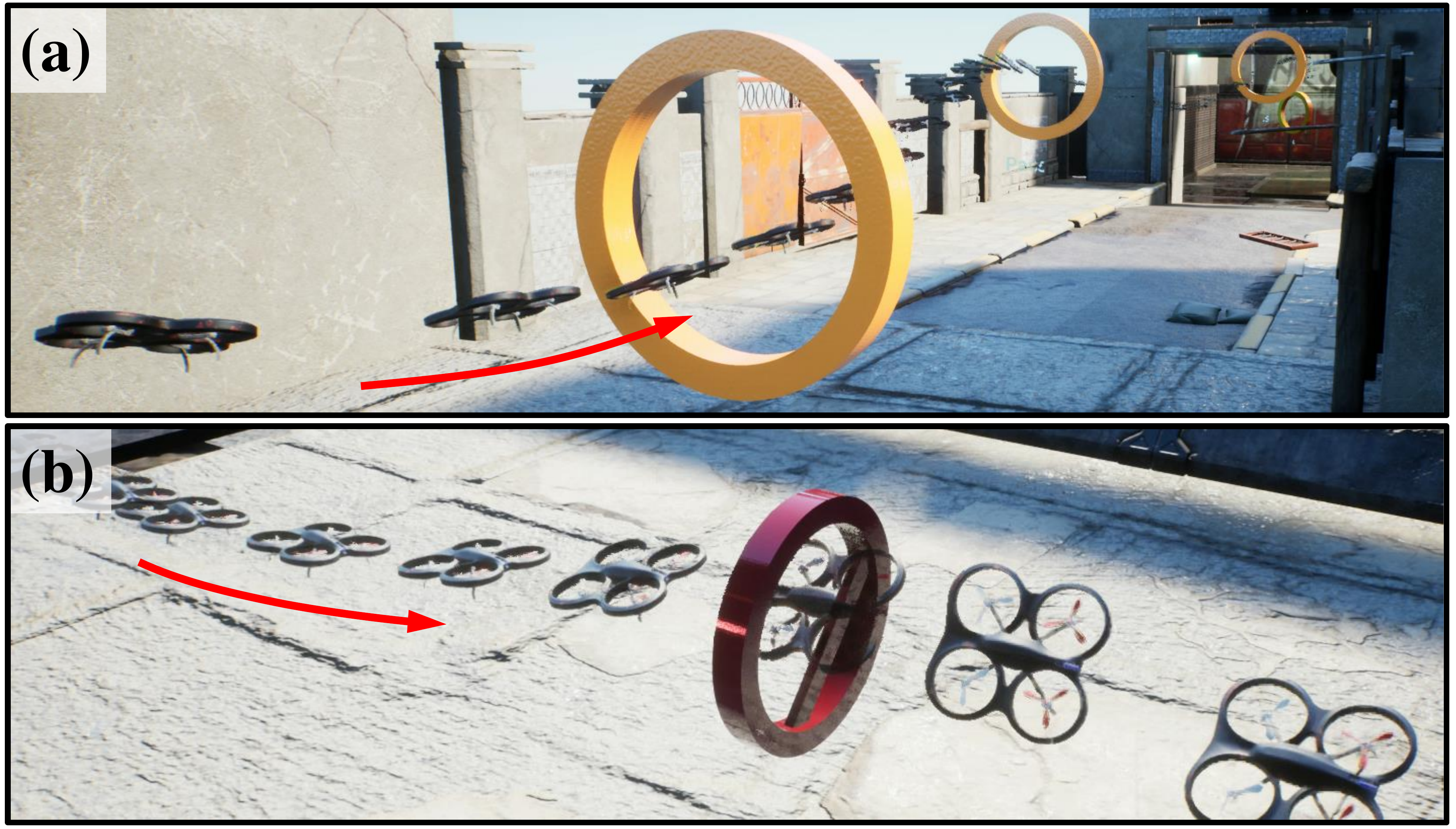}
    \captionsetup{font={footnotesize}}
	\caption{
        Illustration of the flight of our system in the 2022 DJI Robomaster Intelligent UAV Championship.
	}
	\label{fig:bisai}
    \vspace{-0.9cm}
\end{figure}

We take part in the second event Autonomous Racing of this competition, where the quadrotor is required to fly in a dynamic and challenging environment as fast as possible.
The racing track is about $160$ m long,  which contains a series of static gates (as shown in Fig.~\ref{fig:bisai}.\textbf{(a)}), moving obstacles, dynamic gates with different movement patterns (as illustrated in Fig.~\ref{fig:toutu}), and a shaped gate which demands to plan SE(3) trajectory (as shown in Fig.~\ref{fig:bisai}.\textbf{(b)}).
The radius of the gates is $1$ m.
The positions of the gates are provided ahead of each race up to approximately $2$ m uncertainty.
This requires the drone to be able to detect gates and make trajectory adjustments in real time, such as online replanning based on the detection results.

\begin{table}[H]
	\centering
	\caption{Parameters of Planning} 
    \tabcolsep=0.25cm 
	\begin{tabular}{cccccccc}
		\toprule
		$N$ & $M$ &  $m$ & $\lambda_{t}$ & $\lambda_{b}$ & $\lambda_{g}$ & $\lambda_{d}$  \\
		\midrule
		~~2~~ & ~~4~~ & ~~3~~ & ~100~ & 100 & 10000 & 10000   \\
		\bottomrule
	\end{tabular}
	\label{tab:planning}
	\vspace{-0.1cm}
\end{table} 

In the competition, we integrate the proposed framework into a customized quadrotor system, combining localization, detection, and control modules.
For the planning module, the parameters defined in Sec.~\ref{sec:Online_Replan} of trajectory optimization are shown in Tab.~\ref{tab:planning}.
The average time overhead of replan is $16.6$ ms when considering SE(3) for narrow gaps and $7.0$ ms when not considering.
We opt for VINS-MONO~\cite{qin2018vins} as the localization module.
Then we use the point cloud from the depth image for the detection of gates and dynamic obstacles.
Finally, we adopt MPC~\cite{faessler2017differential} as the control method to track the planned aggressive trajectory.

We visualize the flight of our system in the competition in Fig.~\ref{fig:toutu} and \ref{fig:bisai}.
Readers can get a better understanding of the experiment from the attached video.
Additionally, we show the final results and rankings in Tab.~\ref{tab:dji}, where our system is significantly faster than the other teams, demonstrating the great performance of our method for racing in dynamic environments.


\begin{table}[H]
	\centering
	\caption{Ranking of Competition Results\textsuperscript{\ref{foot_rankings}}} 
    \tabcolsep=0.25cm 
	\begin{tabular}{c|cccc}
		\toprule
		rankings & \bf our team & 2nd place & 3rd place & \bf \dots  \\
		\midrule
		completion time (s) & \bf 22.0 &  50.3 &  76.9 & \bf \dots  \\
		\bottomrule
	\end{tabular}
	\vspace{-0.2cm}
	\label{tab:dji}
\end{table} 

\section{Conclusion}

In this paper, we propose a polynomial-based trajectory planning method to address the $4$ requirements for racing in dynamic environments presented in Sec.~\ref{sec:Introduction}.
Efforts in trajectory representation, hard constraint designed for crossing waypoints, and parallel evaluation of trajectory under different topologies, effectively improve replan efficiency, the accuracy of waypoint traversal, and flight time when facing dynamic obstacles.
Finally, the method is applied to the DJI competition, and the outstanding result proves the good performance of our method.

\bibliography{references}
\end{document}